\documentclass[11pt]{article}

\usepackage{times}
\usepackage{coling2016}
\usepackage{caption}
\usepackage{subcaption}
\usepackage{microtype}

\usepackage{url}
\usepackage{pdfsync}
\usepackage{latexsym}
\usepackage{algorithm}
\usepackage{algpseudocode}
\usepackage{graphicx}
\usepackage{fixltx2e}
\usepackage{epstopdf}
\usepackage{amssymb}
\usepackage{amsmath}
\usepackage{xspace,relsize,arydshln,gensymb}
\usepackage{multirow}
\usepackage{graphics,epsfig,boxedminipage}
\usepackage{adjustbox}

\newcommand{\secref}[2][]{Section#1~\ref{sec:#2}}
\newcommand{\figref}[2][]{Figure#1~\ref{fig:#2}}
\newcommand{\tabref}[2][]{Table#1~\ref{tab:#2}}

\newcommand{\dataset}[1]{\textsc{#1}\xspace}
\newcommand{\books}{\dataset{books}}
\newcommand{\news}{\dataset{news}}
\newcommand{\blogs}{\dataset{blogs}}
\newcommand{\pubmed}{\dataset{pubmed}}

\newcommand{\ex}[1]{\textit{#1}\xspace}
\newcommand{\term}[1]{\textit{#1}\xspace}
\newcommand{\topic}[1]{$\langle$\ex{#1}$\rangle$\xspace}
\newcommand{\topiclabel}[1]{\textsc{#1}\xspace}
\newcommand{\bb}[1]{\textbf{#1}}
\newcommand{\diff}{\smaller[2]}

\newcommand\email{\begingroup \urlstyle{tt}\smaller\Url}

\newcommand{\method}[1]{{\smaller[1]\texttt{#1}}\xspace}
\newcommand{\doctovec}{\method{doc2vec}}
\newcommand{\dbow}{\method{dbow}}
\newcommand{\dmpv}{\method{dmpv}}
\newcommand{\wordtovec}{\method{word2vec}}
\newcommand{\sg}{\method{skip-gram}}
\newcommand{\cbow}{\method{cbow}}
\newcommand{\raco}{\method{RACO}}

\newcommand{\lgnb}{LGNB\xspace}
\newcommand{\netl}{NETL\xspace}

\newcommand{\feature}[1]{\textit{#1}\xspace}
\newcommand{\lettertrigram}{\feature{LetterTrigram}}
\newcommand{\pagerank}{\feature{PageRank}}
\newcommand{\wordnum}{\feature{NumWords}}
\newcommand{\topicoverlap}{\feature{TopicOverlap}}

\title{Automatic Labelling of Topics with Neural Embeddings}

\author{Shraey Bhatia$^{1,2}$, Jey Han Lau$^{1,2}$ \and Timothy 
    Baldwin$^{2}$ \\
    $^1$ IBM Research \\
    $^2$ Dept of Computing and Information Systems,\\The University of 
Melbourne \\
    \email{shraeybhatia@gmail.com}, \email{jeyhan.lau@gmail.com}, 
\email{tb@ldwin.net}}

\begin{document}

\maketitle

\begin{abstract}

Topics generated by topic models are typically represented as list of 
terms. To reduce the cognitive overhead of interpreting these topics for 
end-users, we propose labelling a topic with a succinct phrase that 
summarises its theme or idea. Using Wikipedia document titles as label 
candidates, we compute neural embeddings for documents and words to 
select the most relevant labels for topics. Compared to a 
state-of-the-art topic labelling system, our methodology is simpler, 
more efficient, and finds better topic labels.

\end{abstract}

\section{Introduction} 
\label{sec:introduction}

\blfootnote{This work is licensed under a Creative Commons Attribution 
4.0 International License. License details: 
  \url{http://creativecommons.org/licenses/by/4.0/}
}

Topic models are a popular approach to detecting trends and traits in
document collections, e.g.\ in tracing the evolution of a scientific
field through its publications \cite{Hall+:2008}, enabling visual
navigation over search results \cite{Newman+:2009a}, interactively
labelling document collections \cite{Poursabzi-Sangdeh+:2016}, or detecting trends
in text streams \cite{Wang:McCallum:2006,AlSumait+:2008}. They are
typically unsupervised, and generate ``topics'' $t_i$ in the form of
multinominal distributions over the terms $w_j$ of the document
collection ($\Pr(w_j|t_i)$),
and topic distributions for each document $d_k$ in the collection, in the form
of a multinomial distribution over topics ($\Pr(t_i|d_k)$). Traditionally, this has been
carried out based on latent Dirichlet allocation (LDA:
\newcite{Blei+:2003}) or extensions thereof, but more recently there has
been interest in deep learning approaches to topic modelling 
\cite{Cao+:2015,Larochelle+:2012}. 

In contexts where the output of the topic model is presented to a human
user, a fundamental concern is the best way of presenting the rich
information generated by the topic model, in particular, the topics
themselves, which provide the primary insights into the document
collection. The de facto topic representation has been a simple term
list, in the form of the top-10 terms in a given topic, ranked in
descending order of $\Pr(w_j|t_i)$. The cognitive overhead in
interpreting the topic presented as a list of terms can be high, and has
led to interest in the task of generating labels for topics, e.g.\ in
the form of textual descriptions \cite{Mei+:2007,Lau+:2011a,Kou+:2015},
visual representations of the topic words \cite{Smith+:2016-pre}, or
images \cite{Aletras:Stevenson:2013a}. In the former case, for example,
rather than the top-10 terms of \topic{school, student, university,
  college, teacher, class, education, learn, high, program}, a possible
textual label could be simply \topiclabel{education}. Recent work has
shown that, in the context of a timed information retrieval (IR) task,
automatically-generated textual labels are easier for humans to
interpret than the top-10 terms, and lead to equivalent-quality
relevance judgements \cite{Aletras+:2014}. Despite this, the accuracy of
state-of-the-art topic generation methods is far from perfect, providing
the motivation for this work.

In this paper, we propose an approach to topic labelling based on word
and document embeddings, which both automatically generates label
candidates given a topic input, and ranks the candidates in either an
unsupervised or supervised manner, to produce the final topic label. Our
contributions in this work are: (1) a label generation approach based on
combined word and document embeddings, which is both considerably
simpler than and empirically superior to the state-of-the-art generation
method; (2) a simple label ranking approach that exploits character and 
lexical information, which is also superior to the state-of-the-art 
ranking approach; and (3) release of an open source implementation of 
our method, including a new dataset for topic label ranking 
evaluation.\footnote{\url{https://github.com/sb1992/NETL-Automatic-Topic-Labelling-}}

\section{Related Work} 
\label{sec:background}

\newcite{Mei+:2007} introduced the task of generating labels for LDA
topics, based on first extracting bigram collocations from the
topic-modelled document collection using a lexical association measure, and
then ranking them based on KL divergence with each topic. The approach
is completely unsupervised.

\newcite{Lau+:2011a} proposed using English Wikipedia to automatically
label topics. First, they map the topic to a set of concepts by querying
Wikipedia using the top-10 topic terms based on: (a) Wikipedia's native
search API; and (b) Google's search API, with site restriction. The
top-8 article titles from each of these two sources are pooled to
generate the primary candidate topic labels. Secondary labels are 
generated from component $n$-grams contained within the primary
candidates, and filtering out incoherent and unrelated titles using the
\raco measure \cite{Grieser+:2011} to measure similarity with the
primary labels, based on Wikipedia document categories.  The
combined set of primary and secondary label candidates is then ranked
using a number of lexical association features, either directly in an
unsupervised manner, or indirectly based on training a support vector
regression model. The authors provide an extensive analysis of their
method with that of \newcite{Mei+:2007}, and find their label
generation and ranking methodology to be empirically superior (in both an unsupervised
and supervised setting). In this paper, we seek to improve upon the
topic labelling benchmark set by \newcite{Lau+:2011a}.

\newcite{hulpus+:2013} developed a graph-based method for topic
labelling, leveraging the structure of DBpedia concepts. Their approach
is styled around graph-based word sense disambiguation, and extracts a
set of DBpedia concepts corresponding to the top-$N$ terms of a topic.
They then construct a graph centred around DBpedia concepts and
filter noise based on graph connectivity (the hypothesis being that
sense graphs of words from a topic should be connected). To find the
best label for a topic, they experiment with a variety of graph centrality measures.

In work slightly further afield, \newcite{zhao+:2011} proposed topical
keyphrase extraction for Twitter.  Although the work focuses mainly on
Twitter, the methodology can be applied to other domains and to label
topics.  \newcite{zhao+:2011} follow a three-step process for keyphrase
extraction: (1) keyword ranking; (2) candidate keyphrase generation
(based on the individual keywords); and (3) keyphrase ranking. They use
a novel topic context-sensitive \pagerank method to regularise topic
scores for keyword ranking, and a probabilistic scoring method that
takes into account relevance, interestingness and keyphrase length for
keyphrase ranking. 

Not directly for the purposes of topic labelling but as a relevant
pretraining method, \newcite{Mikolov+:2013} proposed \wordtovec to
learn word embeddings, which they found to perform strongly over a range
of word similarity tasks, and also to be useful for initialising deep
learning models. Two approaches are proposed in the paper: \cbow and
\sg. \cbow combines neighbouring words to predict a target word, while
\sg uses the target word to predict neighbouring words.  Both approaches
use a feedforward neural network with a non-linear hidden layer to
maximize the objective function; to improve computational efficiency,
the authors propose using negative sampling.  In this paper, we will use
\wordtovec as a means of generating topic term and label
representations.

As an extension of \wordtovec, \newcite{Le+:2014} introduced \doctovec  
to learn embeddings for word sequences (e.g.\ paragraphs or documents).  
By treating each document as a word token, the same \wordtovec 
methodology can be used to learn document embeddings. The authors 
propose two implementations: \dbow, which uses the document vector to 
predict its document words, and is the \doctovec equivalent of \sg; and 
\dmpv, which uses a small window of words and concatenates them with the 
document vector to predict a document word, and is the \doctovec 
equivalent of \cbow.\footnote{Strictly speaking, \cbow combines word 
vectors by summing them, while \dmpv combine word vectors and document 
vector by concatenating them.}  Compared to \dbow, \dmpv takes into 
account the local word ordering, and has a higher number of parameters, 
since the input is a concatenation of vectors. As with \wordtovec, we 
will use \doctovec as an alternative means of generating topic term and 
label representations.

Building on \wordtovec, \newcite{Kou+:2015} experimented with neural
embeddings in the context of topic labelling.  In addition to \sg and
\cbow word vectors, the authors also included letter trigram vectors of
a word, with the rationale that it generalises over
morphologically-related forms of the same word.  Their methodology
consists of first generating candidate labels for topics from
topic-related documents using a chunk parser. By representing both topic
words and topic labels using word embeddings and letter trigrams, they
rank the labels using cosine similarity to obtain the best label for a
topic. In their evaluation, they find that simple letter trigrams are
ultimately the most reliable means of label ranking.

\section{Methodology} 
\label{sec:methodology}

Following \newcite{Lau+:2011a}, our method is made up of two steps: (1)
topic label generation based on English Wikipedia; and (2) topic label
ranking, based on a supervised learn-to-rank model. We detail each of
these steps below.

\subsection{Candidate Generation}
\label{sec:cand-gen}

To match topics to Wikipedia articles,\footnote{As of 2016 there are
  over 5 million documents in English Wikipedia.}  \newcite{Lau+:2011a}
used an IR approach, by querying English Wikipedia with the top-$N$ topic
terms. 
However, in order to do this, they required external resources (two
search APIs, one of which is no longer publicly available), limiting the
general-purpose utility of the method. We propose an alternative
approach: precomputing distributed representations of the topic terms
and article titles using \wordtovec and \doctovec.

To this end, we train a \doctovec model on the English
Wikipedia articles, and represent the embedding of a Wikipedia title by 
its document embedding.
As \doctovec runs \wordtovec internally, word embeddings are also learnt
during the training.  Given the top-$N$ topic terms, the topic embedding 
is represented by these terms' word embeddings.  Based on the findings 
of \newcite{Lau:Baldwin:2016b} that the simpler \dbow has less
parameters, trains faster, and performs better than \dmpv in several
extrinsic tasks, we experiment only with \dbow.\footnote{We use Gensim's
  implementation of both \doctovec and \wordtovec for all experiments:
  \url{https://radimrehurek.com/gensim/}.}  In terms of hyper-parameter
settings, we follow the recommendations of 
\newcite{Lau:Baldwin:2016b}.\footnote{\doctovec hyper-parameters: 
sub-sampling threshold $= 10^{-5}$, vector size $= 300$, window size $= 
15$, negative sample size $= 5$, and training epochs $= 20$.}

In addition to \doctovec, we also experiment with \wordtovec to
generate embeddings for Wikipedia titles. By treating titles as
a single token (e.g.\ concatenating \term{financial crisis} into
\term{financial\_crisis}) and greedily tokenising the text of all of the
Wikipedia articles, we can then generate word embeddings for the
titles. For \wordtovec, we use the \sg implementation
exclusively.\footnote{\wordtovec hyper-parameters: sub-sampling 
threshold $= 10^{-5}$, vector size $= 300$, window size $= 5$, negative 
sample size $= 5$, and training epochs $= 100$.}

For both \doctovec and \wordtovec, we first pre-process English
Wikipedia,\footnote{The English Wikipedia dump used in all experiments
  is dated 2015-12-01.}  using Wiki Extractor to clean and extract
Wikipedia articles from the original
dump.\footnote{\url{https://github.com/attardi/wikiextractor/}} We then
tokenise words with the Stanford CoreNLP Parser
\cite{Klein:Manning:2003}, and lowercase all words.  We additionally 
filter out articles where the article
body is made up of less than 40 words, and also disambiguation pages. We
also remove titles whose length is longer than 4 words, as they are
often too specific or inappropriate as topic labels (e.g.\ \term{List of
  Presidents of the United States}). For \wordtovec, we remove any parenthesised 
sub-component of an article title --- e.g.\ in the 
case of \term{Democratic Party (United States)}, we remove \term{(United States)} to generate 
\term{Democratic Party} --- as we would not expect to find verbatim usages 
of the full title.  This has the potential side-effect of mapping 
multiple articles onto a single ambiguous title, resulting in multiple 
representations for \term{Democratic Party}. While acknowledging that 
there are instances where the more specific title may be appropriate as 
a label, the generalised version is always going to be a hypernym of the 
original, and thus appropriate as a label candidate.


Given a topic, we measure the relevance of each title embedding
(generated by either \doctovec or \wordtovec) based on the pairwise
cosine similarity with each of the word embeddings for the top-10
topic terms, and aggregate by taking the arithmetic mean. Formally, the
\doctovec relevance ($rel_{d2v}$) and \wordtovec relevance ($rel_{w2v}$)
of a title $a$ and a topic $T$ is given as follows:
\begin{align}
\label{eqn:doctovec}
rel_{d2v}(a, T) &= \frac{1}{|T|} \sum_{v \in T} \cos \Big( 
E^{d}_{d2v}(a), E^{w}_{d2v}(v) \Big) \\
\label{eqn:wordtovec}
rel_{w2v}(a, T) &= \frac{1}{|T|} \sum_{v \in T} \cos \Big( 
E^{w}_{w2v}(a), E^{w}_{w2v}(v) \Big)
\end{align}
where $E_{d2v}^{d}(x)$ is the document embedding of title $x$ generated
by \doctovec; $E_{d2v}^{w}(y)$ is the word embedding of word $y$ generated
by \doctovec;
$E_{w2v}^{w}(z)$ is the word embedding of word $z$ generated by
\wordtovec; $v\in T$ is a
topic term; $|T|$ is the number of topic terms (10 in our experiments);
and $\cos(\vec{x},\vec{y})$ is the cosine similarity function.


The idea behind using both \doctovec and \wordtovec to generate title
embeddings is that we observe that the two models favour different types of
labels: \doctovec tends to favour fine-grained concepts, while \wordtovec
favours more generic or abstract labels. As an illustration of this, in
\tabref{examples} we present one of the actual topics used later in our
evaluation, and the top-5 article titles based on \doctovec and
\wordtovec. This dichotomy is rooted in the differences in the modelling
of context in the two models. In \doctovec, the title embedding is
determined by the words that belong to the title, each of which is in
turn determined by its context of use; it thus directly captures the
compositional semantics of the title. With our \wordtovec method, on the
other hand, the title embedding is determined directly by the
neighbouring words of the title token in text, oblivious to the
composition of words in the title. 

\begin{table}[t]
\centering
\begin{tabular}{lcc}
\textbf{Top-10 Topic Terms} & \textbf{\wordtovec Labels} & \textbf{\doctovec 
Labels} \\
\hline
& software & microsoft\ visual\ studio \\
blogs, vmware, server, virtual, oracle, & desktop & 
desktop\ virtualization \\
update,  virtualization,  application, & operating\ system & 
microsoft\ exchange\ server \\
infrastructure,  management & virtualization & cloud\ computing \\
& middleware & windows\ server\ 2008 \\
\end{tabular}

\caption{The top-5 labels generated using \doctovec and \wordtovec title 
embeddings for the topic provided}
\label{tab:examples}
\end{table}

To combine the strengths of \doctovec and \wordtovec, for each topic we
generate a combined candidate ranking by summing the relevance scores 
using top-100 candidates from \doctovec and \wordtovec:\footnote{From 
preliminary experiments we found that summing only the top-100 
candidates from \doctovec and \wordtovec is better than summing all 
candidates. As we remove the parenthesised sub-component of an article 
title for \wordtovec (\term{Democratic Party (United States)} 
$\longrightarrow$ \term{Democratic Party}) , we observe that these 
titles tend to be very general and can ocassionally produce very high 
cosine similarity and skew the combined score for a number of similar 
labels (e.g. causing \term{Democratic Party} from a host of countries 
(\term{Democratic Party (United States)}, \term{Democratic Party 
(Australia)}, etc) to appear in the top ranking).}
\begin{equation}
\label{eqn:docwordtovec}
rel_{d2v+w2v}(a, T) = rel_{d2v}(a,T) + rel_{w2v}(a,T)
\end{equation}

\subsection{Candidate Ranking}
\label{sec:cand-rank}

The next step after candidate generation is to re-rank them based on a supervised learn-to-rank
model, in an attempt to improve the quality of the top-ranking
candidates.

The first feature used in the supervised reranker is \lettertrigram, and
based on the finding of \newcite{Kou+:2015} that letter trigram vectors
are an effective means of ranking topic labels.  Our implementation of
their method is based on measuring the overlap of letter trigrams
between a given topic label and the topic words. For each topic, we
first convert each topic label and topic words into multinomial
distributions over letter trigrams, based on simple maximum likelihood
estimation.\footnote{For topic words, the letter trigrams are generated
  by parsing each of the topic words as separate strings rather than one
  concatenated string.}  We then rank the labels based on their cosine
similarity with the topic words.  The rank value constitutes the first
feature of the supervised learn-to-rank model.  Additionally, this
ranking by letter trigram method also forms our unsupervised baseline,
as we found that it to have the best unsupervised ranking performance of
all our features, consistent with the findings of
\newcite{Kou+:2015}.\footnote{Note that we do not make use of the noun
  chunk-based label generation methodology of \newcite{Kou+:2015}, in
  line with the findings of \newcite{Lau+:2011a} that Wikipedia titles
  give rise to better label candidates than $n$-grams extracted from the
  topic-modelled documents.}

The second feature is \pagerank \cite{Page+:1998}, in an attempt to
prefer labels which represent more ``core'' concepts in Wikipedia.
\pagerank uses directed links to estimate the significance of a
document, based on the probability of a random web surfer visiting a web
page by either following hyperlinks or randomly transporting to a new
page.  We construct a directed graph from Wikipedia based on hyperlinks
within the article text, and from this, compute a \pagerank value for
each Wikipedia article (and hence, title).\footnote{We use the following
  implementation for \pagerank:
  \url{https://www.nayuki.io/page/computing-wikipedias-internal-pageranks}}

Our last two features are lexical features proposed by
\newcite{Lau+:2011a}: (1) \wordnum, which is simply the number of words in
the candidate label (e.g.\ \term{operating system} has 2 words); and (2)
\topicoverlap, which is the lexical overlap between the candidate label
and the top-10 topic terms (e.g.\ \term{desktop virtualization} has a
\topicoverlap score of 1 in our example from \tabref{examples}).

Given these features and a gold standard order of candidates (detailed 
in \secref{cand-lab}), we train a support vector regression model (SVR: 
\newcite{Joachims:2006}) over these four features. 

\section{Datasets} 
\label{sec:datasets}

For direct comparison with \newcite{Lau+:2011a}, we use the same set of
topics they used in their experiments. These were generated from 4
different domains: \blogs, \books, \news and \pubmed.  In general,
\blogs, \books and \news cover wide-ranging topics from product reviews
to religion to finance and entertainment, whereas PubMed is
medical-domain specific.

\blogs is made up of 120k blog articles from the Spinn3r blog dataset; \books 
is made up of 1k English language books from the Internet Archive American 
Libraries collection; \news is made up of 29k New York Times articles from 
English Gigaword; and \pubmed is made up of 77k PubMed biomedical 
abstracts.  \newcite{Lau+:2011a} ran LDA on these documents and 
generated 100 topics for each domain. They filtered incoherent topics 
using an automated approach \cite{Newman+:2010a}, resulting in 45, 38, 60, 85 topics for 
\blogs, \books, \news and \pubmed, respectively.

\subsection{Gold Standard Judgements}
\label{sec:cand-lab}

To evaluate our method and train the supervised model, gold-standard
ratings of the candidates are required. To this end, we used CrowdFlower
to collect human
judgements.\footnote{\url{https://www.crowdflower.com/}}
We follow the approach of \newcite{Lau+:2011a}, presenting 10 pairings
of topic and candidate label, and asking human judges to rate the label on an 
ordinal scale of 0--3 where 0 indicates a completely inappropriate
label, and 3 indicates a very good label for the given topic.

To control for annotation quality, we make use of the original
annotations released by \newcite{Lau+:2011a}.  We select labels with a
mean rating $\geq 2.5$ (good labels) and $\leq 0.5$ (bad labels) to
serve as controls in our tasks.  We include an additional topic--label
control pair in addition to the 10 topic--label pairs in a HIT. Control
pairs are selected randomly without replacement, and randomly injected
into the HIT.  To pass quality control, a worker is required to rate bad
labels $\leq 1.0$ and good labels $\geq 2.0$. A worker is filtered out 
if his/her overall pass rate over all control pairs is $< 0.75$.

Each candidate label was rated by 10 annotators. Post-filtered, we have
an average of 6.4 annotations for each candidate
label.\footnote{Post-filering, some candidates ended up with less than 3
  annotations; these candidates were posted for another annotation round
  to gather more annotations.} To aggregate the ratings for a candidate
label, we compute its mean rating, and rank the candidate labels
based on the mean ratings to produce the gold standard ranking for each
topic.

We collect judgements for the top-19 candidates from the unsupervised
ranking.\footnote{Ideally we would have liked to have collected
  judgements for as many candidates as possible, but due to budget
  constraints we were only able to have the top-19 annotated.} For
candidate ranking (\secref{cand-rank}), we are therefore re-ranking the
top-19 candidates.

\section{Experiments} 
\label{sec:experiments}

In this section we present the results of our topic labelling
experiments, and compare our method with that of
\newcite{Lau+:2011a}. Henceforth we refer to our method as ``\netl'' (neural
embedding topic labelling), and \newcite{Lau+:2011a} as ``\lgnb''.

Following \lgnb, we use {\bf top-1 average rating} and {\bf normalized
  discounted cumulative gain (nDCG)}
\cite{Jarvelin:Kekalainen:2002,Croft+:2009} as our evaluation metrics.
Top-1 average computes the mean rating of the top-ranked labels, and
provides an evaluation of the absolute utility of the preferred
labels. nDCG, on the other hand, measures the relative quality of the
ranking, calibrated relative to the ratings of the gold-standard
ranking. Similarly to \lgnb, we compute nDCG for the top-1 (nDCG-1), top-3
(nDCG-3), and top-5 (nDCG-5) ranked labels.

\subsection{Results}
\label{sec:results-supervised}

\begin{table}[t!]
\centering
\begin{adjustbox}{max width=\textwidth}
\begin{tabular}{clc@{\;}cc@{\;}cc@{\;}cc@{\;}c}
\multirow{2}{*}{\textbf{Test Domain}} & 
\multirow{2}{*}{\textbf{Training}} & \multicolumn{2}{c}{\textbf{Top-1 
Avg.}} & \multicolumn{2}{c}{\textbf{nDCG-1}} & 
\multicolumn{2}{c}{\textbf{nDCG-3}} & 
\multicolumn{2}{c}{\textbf{nDCG-5}} \\
 &  & \textbf{\lgnb} & \textbf{\netl} & \textbf{\lgnb} & \textbf{\netl} & \textbf{\lgnb} & \textbf{\netl} & \textbf{\lgnb} & \textbf{\netl} \\
 \hline
\multirow{6}{*}{\textbf{\blogs}}
 & Baseline & 1.84 & \bb{1.91} & 0.75 & \bb{0.77} & 0.77 & \bb{0.82} & 0.79 & 
\bb{0.83} \\
\cdashline{2-10}
 & In-Domain              & 1.98 & \bb{2.00} & 0.81 & 0.81 & 0.82 & \bb{0.85} & 0.83 & 0.84 \\
 & Cross-domain: \books   & 1.88 & \bb{1.91} & 0.77 & 0.78 & 0.81 &\bb{0.83} & 0.83 & 0.83 \\
 & Cross-domain: \news    & \bb{1.97} & 1.92 & \bb{0.80} & 0.78 & 0.83 & 0.84 & 0.83 & 0.84 \\
 & Cross-domain: \pubmed  & \bb{1.95} & 1.90 & \bb{0.80} & 0.77 & 0.82 & 0.83 & 0.83 & 0.83 \\
 & Cross-domain: All 3    & --- & 1.92 & --- & 0.78 & --- & 0.84 & --- & 0.84 \\ 
\cdashline{2-10}
 & Upper Bound            & 2.45 & \bb{2.48}  & 1.00 & 1.00 & 1.00 & 1.00 & 1.00 & 1.00 \\
 \hline
\multirow{6}{*}{\textbf{\books}}
 & Baseline & {1.75} & \bb{1.97} & 0.77 & 0.78 & 0.77 &\bb{ 0.82} & 
0.79 & \bb{0.83} \\
\cdashline{2-10}
 & In-Domain              & 1.91 & \bb{1.99} & \bb{0.84} & 0.82 & 0.81 & 0.82 & 0.83 & 0.84 \\
 & Cross-domain: \blogs    & 1.82 & \bb{2.02} & 0.79 &\bb{ 0.83} & 0.81 & 0.82 & 0.82 & \bb{0.84} \\
 & Cross-domain: \news     & 1.82 & \bb{1.99} & 0.79 &\bb{ 0.81} & 0.81 & 0.82 & 0.83 & 0.84 \\
 & Cross-domain: \pubmed   & 1.87 & \bb{1.97} & 0.81 & 0.80 & 0.82 & 0.82 & 0.83 & 0.84 \\
 & Cross-domain: All 3     & --- & 2.03 & --- & 0.83 & --- & 0.83 & --- & 0.84 \\
\cdashline{2-10}
 & Upper Bound            & 2.29 & \bb{2.49} & 1.00 & 1.00 & 1.00 & 1.00 & 1.00 & 1.00 \\
  \hline
\multirow{6}{*}{\textbf{\news}}
 & Baseline & 1.96 & \bb{2.04} & 0.80 & \bb{0.82} & 0.79 & \bb{0.84} & 
0.78 & \bb{0.85} \\
\cdashline{2-10}
 & In-Domain              & 2.02 & 2.02  & \bb{0.82} & 0.80 & 0.82 & \bb{0.84} & 0.84 & 0.85 \\
 & Cross-domain: \blogs    & 2.03 & 2.03 & \bb{0.83} & 0.81 & 0.82 & \bb{0.84} & 0.84 & 0.85 \\
 & Cross-domain: \books    & \bb{2.01} & 1.98 & \bb{0.82} & 0.79 & 0.82 & 0.83 & 0.83 & 0.84 \\
 & Cross-domain: \pubmed   & 2.01 & 2.00 & \bb{0.82} & 0.79 & 0.82 & 0.83 & 0.83 & 0.84 \\
 & Cross-domain: All 3     & --- & 1.99 & --- & 0.79 & --- & 0.84 & --- & 0.84 \\ 
\cdashline{2-10}
 & Upper Bound            & 2.45 & \bb{2.56} & 1.00 & 1.00 & 1.00 & 1.00 & 1.00 & 1.00 \\
  \hline
\multirow{6}{*}{\textbf{\pubmed}}
 & Baseline & 1.73 & \bb{1.94} & 0.75 & \bb{0.79} & 0.77 & \bb{0.80} & 0.79 & \bb{0.82} 
\\
\cdashline{2-10}
 & In-Domain              & 1.79 & \bb{1.99} & 0.77 &\bb{ 0.81} & 0.82 & 0.81 & \bb{0.84} & 0.82 \\
 & Cross-domain: \blogs    & 1.80 & \bb{1.98} & 0.78 & \bb{0.80} & 0.82 & 0.81 & \bb{0.84} & 0.82 \\
 & Cross-domain: \books    & 1.77 & \bb{1.98} & 0.77 & \bb{0.80} & 0.82 & 0.81 & 0.83 & 0.82 \\
 & Cross-domain: \news     & 1.79 & \bb{1.98} & 0.77 & \bb{0.80} & 0.82 & 0.81 & \bb{0.84} & 0.82 \\
 & Cross-domain: All 3     & --- & 2.01 & --- & 0.81 & --- & 0.81 & --- & 0.82 \\
\cdashline{2-10}
 & Upper Bound            & 2.31 & \bb{2.51} & 1.00 & 1.00 & 1.00 & 1.00 & 1.00 & 1.00 \\
 
\end{tabular}
\end{adjustbox}
\caption{Results across the four domains. Boldface indicates the better 
system between \netl and \lgnb (with an absolute difference $> 0.01$).}
\label{tab:sup-results}
\end{table}

Following \lgnb, we present results for: (a) the unsupervised ranker 
(based on letter trigrams); (b) the supervised re-ranker in-domain, 
based on 10-fold cross validation, averaged over 10 runs with different 
partitionings; (c) the supervised re-ranker cross-domain; and (d) the 
upper bound, based on a perfect ranking of the candidates.  For 
cross-domain learning, we train our model using one domain and test it 
on a different domain, or alternatively combine data from three domains 
and test on the remaining fourth domain, e.g.\ training on 
\books$+$\news$+$\pubmed and testing on \blogs. Cross-domain results 
give us a more accurate picture of the performance of our methodology in 
real-world applications (where it would be unrealistic to expect that 
there would be manual annotations of label candidates for that domain). 
We primarily use the in-domain results to gauge the relative quality of 
the cross-domain results.

\begin{table}[t]
\centering
\begin{tabular}{lll}
\textbf{Domain} & \textbf{Topic Terms} & \textbf{Label Candidate} \\
\hline
\blogs & \begin{tabular}[c]{@{}l@{}}vmware server virtual oracle update virtualisation\\  application infrastructure management microsoft\end{tabular} & virtualisation \\
\hdashline
\books & \begin{tabular}[c]{@{}l@{}}church archway building window gothic\\ nave side value tower\end{tabular} & church architecture \\
\hdashline
\news & \begin{tabular}[c]{@{}l@{}}investigation fbi official department federal agent\\ investigator charge attorney evidence\end{tabular} & criminal investigation \\
\hdashline
\pubmed & \begin{tabular}[c]{@{}l@{}}rate population prevalence study incidence datum\\ increase mortality age death\end{tabular} & mortality rate \\
\end{tabular}
\caption{A sample of topics and their topic labels generated by \netl.}
\label{tab:label-example}
\end{table}

We present the results in \tabref{sup-results}, displaying the
performance of \netl and \lgnb side by side for ease of
comparison.\footnote{\lgnb results are taken directly from the original
  paper.} For each domain, the unsupervised baseline of \netl is based 
on the overlap of letter trigrams of the generated candidates with topic 
words (see \secref{cand-rank}). The unsupervised baseline of \lgnb ranks 
the labels using a lexical association measure (Pearson's $\chi^2$). 

Looking at the performance of our method, we can see that the supervised
system improves over the unsupervised baseline across 
all domains with the exception of a small drop observed in \news. 
Surprisingly, there is relatively little difference between 
the in-domain and cross-domain results for our method (but greater 
disparity for \lgnb, especially over \books; for \news, our cross-domain 
models actually outperform the in-domain model). The most consistent 
cross-domain results are generated when we combine all 3 domains, an unsurprising
result given that it has access to the most training data, but
encouraging in terms of having a single model which performs
consistently across a range of domains. 


We next compare \netl to \lgnb, first focusing on the top-1 average
rating metric. The most striking difference is the large improvement
over \pubmed.  \lgnb attributed the poor performance over \pubmed to it
being more domain-specific (and a poorer fit to Wikipedia) than the
other domains, and suggested the need of biomedical experts for
annotation.  Our experiments found, however, that the performance of
\pubmed is comparable to other domains. Additionally, the improvement 
in \books is also quite substantial.   Overall, \netl is more consistent 
across different domains and outperforms \lgnb over
2 domains (\books and \pubmed), and the difference between
\netl and \lgnb is small for \news and \blogs .  The other observation 
is the upper bound performance of \netl is uniformly better than that of 
\lgnb, implying we are also generating better label candidates (we 
revisit this
in detail in \secref{cand-gen-results}).

Moving to nDCG, the performance difference for nDCG-3/5 is largely
indistinguishable for the two systems. \lgnb, however, outperforms \netl
in \news for nDCG-1 whereas \netl does better in\pubmed for nDCG-1 . 


To give a sense of the sort of labels generated by \netl, we present a 
few topics and their top-ranked labels in \tabref{label-example}.


\subsection{Breaking Down \netl vs.\ \lgnb}
\label{sec:compare-lkdt}

The results for \netl and \lgnb in \tabref{sup-results} conflate the
candidate label selection and ranking steps, making it hard to get a
sense of the relative impact of the different design choices implicit in
the two sub-tasks.  To provide a better comparison between the two
methodologies, we present experiments evaluating the candidate
generation and ranking method of the two systems separately.

\subsubsection{Candidate Generation}
\label{sec:cand-gen-results}

In \tabref{sup-results} we saw that \netl has a higher upper bound than
that of \lgnb, suggesting that the generated candidate labels were on
average better.  This is despite the average number of topic label
candidates actually being higher for \lgnb (25 vs.\ 19). 
Here, we present a more rigorous evaluation of the candidate generation
method of both systems.

\begin{figure}[t]
\begin{subfigure}{.3\textwidth}
\centering
\includegraphics[width=\textwidth]{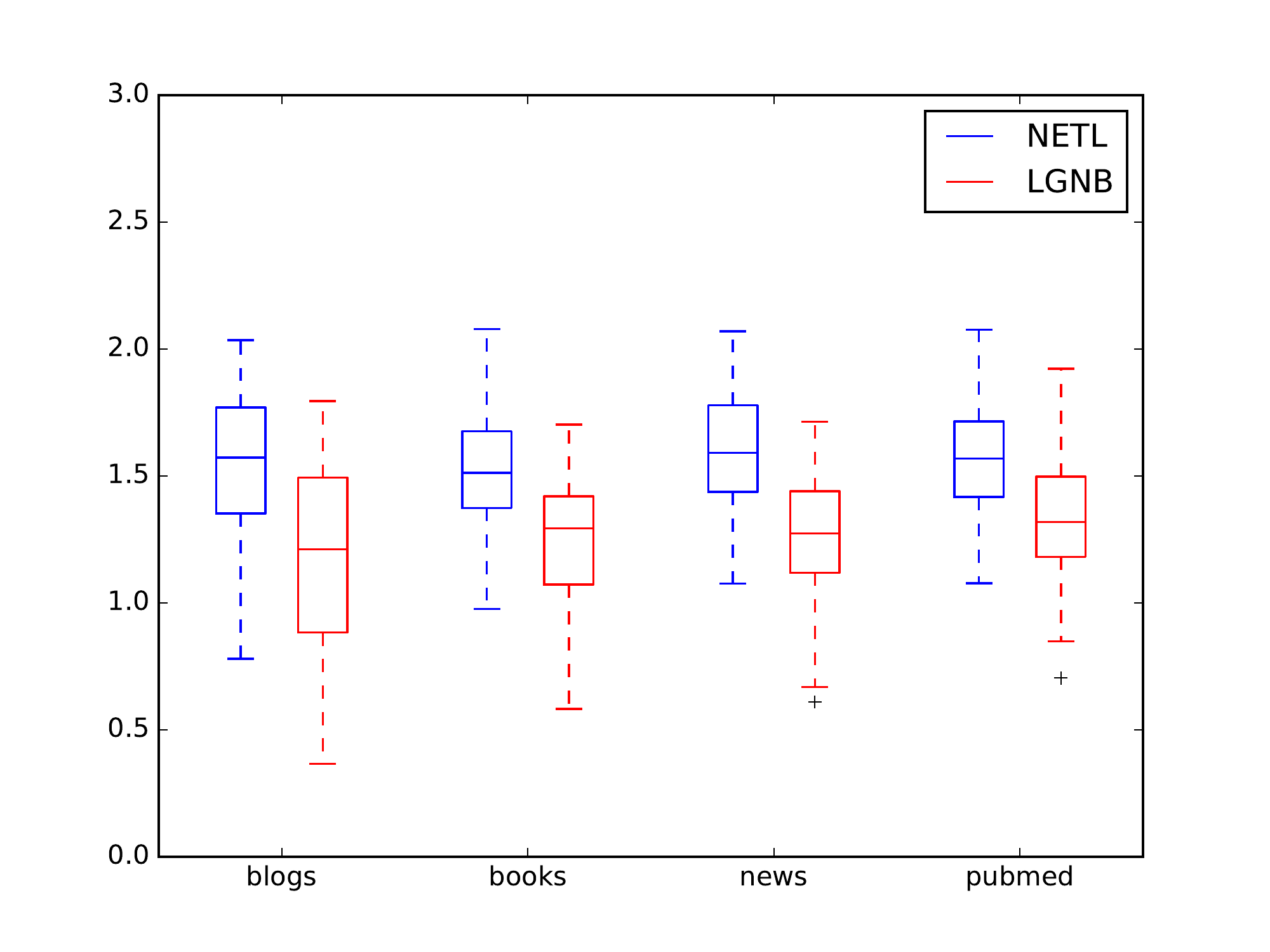}
\caption{Mean Ratings}
\label{fig:mean}
\end{subfigure}
~
\begin{subfigure}{.3\textwidth}
\centering
\includegraphics[width=\textwidth]{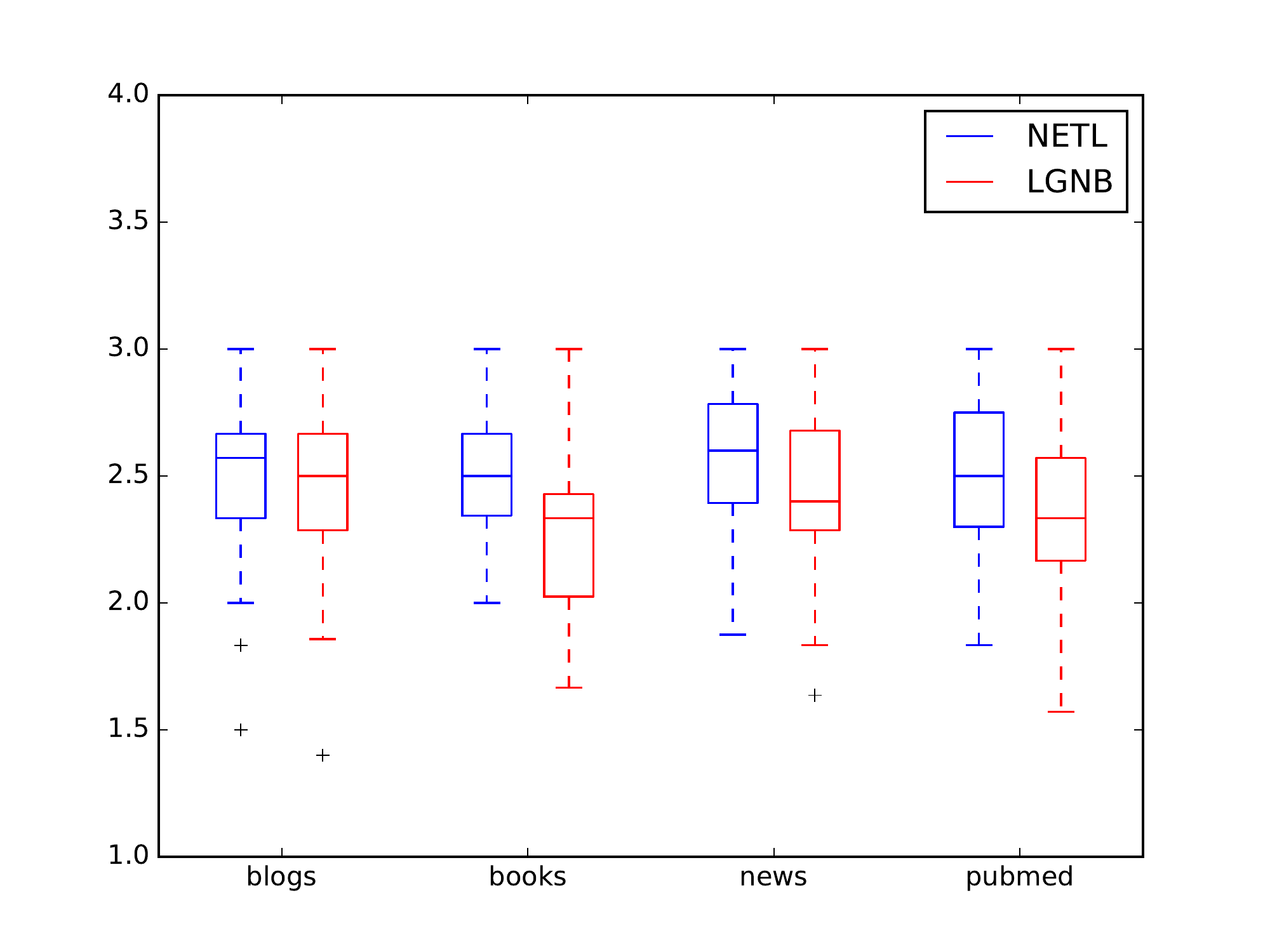}
\caption{Maximum Ratings}
\label{fig:ubound}
\end{subfigure}
~
\begin{subfigure}{.3\textwidth}
\centering
\includegraphics[width=\textwidth]{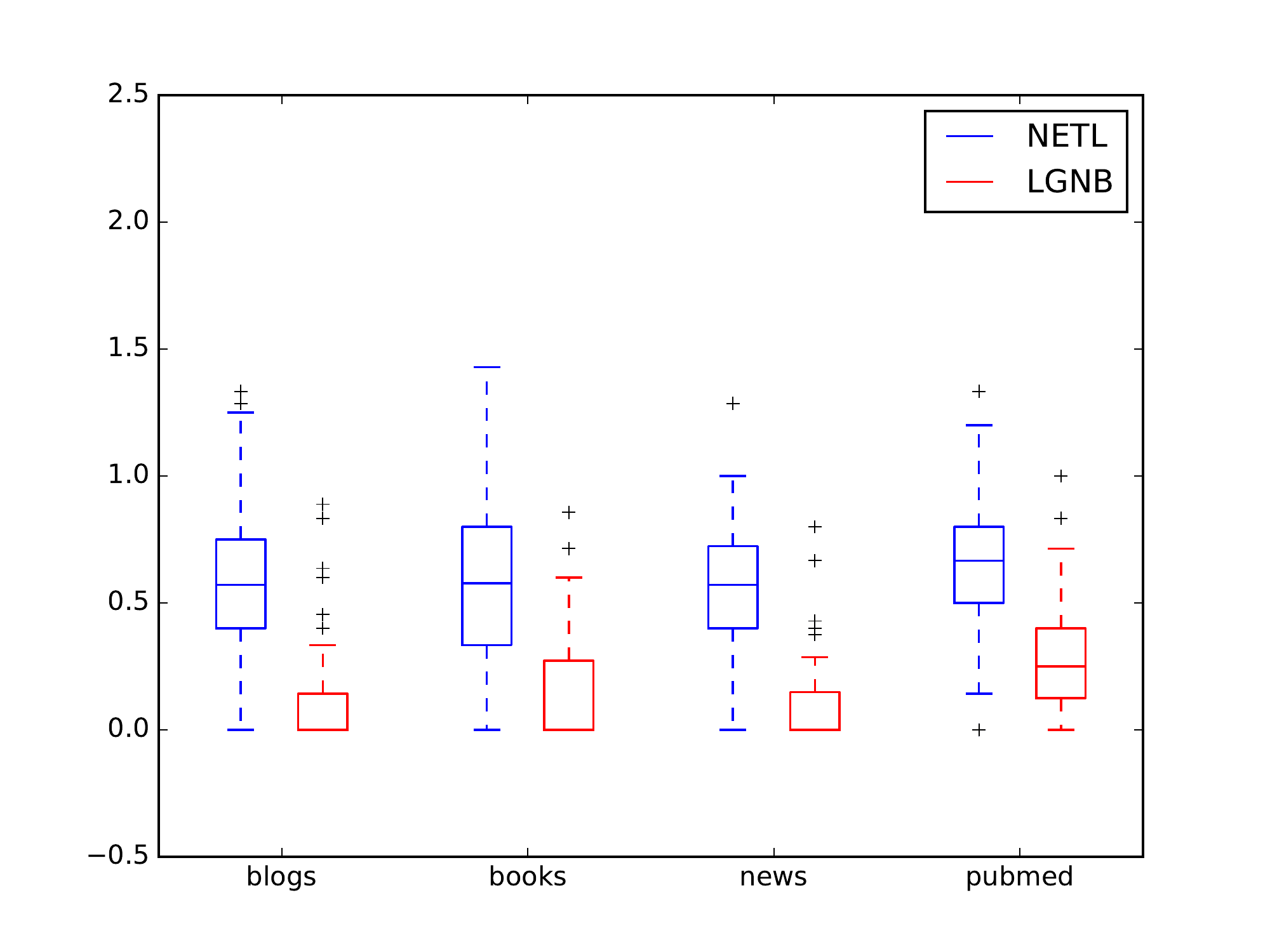}
\caption{Minimum Ratings}
\label{fig:lbound}
\end{subfigure}
\caption{Boxplots of ratings of candidates generated by \netl and 
\lgnb.}
\label{fig:cand-gen-boxplot}
\end{figure}

For each topic, we determine the mean, maximum and minimum label ratings
for a given topic, and plot them in boxplots in
\figref{cand-gen-boxplot}, aggregated per domain. The mean rating
boxplot shows the average quality of candidates, while the maximum
(minimum) rating boxplot reveals the average best (worst) quality of
candidates that are generated by the two systems.

Looking at the boxplots, we see very clearly that \netl generates on
average higher-quality candidates. Across all domains for mean, maximum
and minimum ratings, the difference is substantial.

To provide a quantitative evaluation, we conduct one-sided paired
$t$-tests to test the difference for all pairs in the boxplots. Except
for the maximum ratings on \blogs, all tests are significant
($p < 0.05$). 
These results demonstrate that \netl generates better candidates than
\lgnb (in all of the best-case, average-case and worst-case scenarios).

\subsubsection{Candidate Ranking}
\label{sec:comparison-cand-rank}

Next, we directly compare the ranking method of \netl and \lgnb. Using
candidates generated by \netl, we re-rank the candidates using the
ranking method of each of \lgnb and \netl, and compare the results.

Both \lgnb and \netl train an SVR re-ranker, using a
partially-overlapping set of features. For \lgnb, the ranking
methodology uses 7 lexical association measures (PMI, Student's
$t$-test, Dice's coefficient, Pearson's $\chi^2$ test, log likelihood
ratio, conditional and reverse conditional probability), 2 lexical
features (the same 2 features that \netl uses: \wordnum and
\topicoverlap), and a search engine score based on Zettair. \netl, on
the other hand, uses only 4 features: \lettertrigram, \pagerank, \topicoverlap and \wordnum.

For \lgnb, we exclude the Zettair search engine score feature (as it was 
found to be an unimportant feature), and generate the lexical 
association features by parsing English Wikipedia. We train 2 SVR models 
using \lgnb and \netl features.  Results are presented in 
\tabref{cand-rank-results}.

Using the same candidates, we see that \netl's features produce  
better rankings, outperforming \lgnb's features across all domains.  This 
shows that not only does \netl generate better candidates, but also 
ranks them better than \lgnb.



\begin{table}[t]
\footnotesize
\centering
\begin{tabular}{clcccc}
\multicolumn{1}{l}{\textbf{Test Domain}} & \textbf{Features} & 
\textbf{Top-1 Avg.} & \textbf{nDCG-1} & \textbf{nDCG-3} & 
\textbf{nDCG-5} \\
\hline
\multirow{2}{*}{\textbf{\blogs}} & \lgnb& 1.92 & 0.79 & 0.81 & 0.82 \\
 & \netl & {\textbf{2.00}} & {\textbf{0.81}} & {\textbf{0.85}} & {\textbf{0.84}} \\
 \hdashline
\multirow{2}{*}{\textbf{\books}} & \lgnb & 1.86 & 0.77 & 0.79 & 0.80 \\
 & \netl & {\textbf{1.99}} & {\textbf{0.82}} & {\textbf{0.82}} &{\textbf{0.84}} \\
 \hdashline
\multirow{2}{*}{\textbf{\news}} & \lgnb & 1.87 & 0.75 & 0.79 & 0.81 \\
 & \netl & {\textbf{2.02}} & {\textbf{0.80}} & {\textbf{0.84}} & {\textbf{0.85}} \\
 \hdashline
\multirow{2}{*}{\textbf{\pubmed}} & \lgnb  & 1.89 & 0.77 & 0.79 & 0.81 
 \\
 & \netl  & {\textbf{1.99}} & {\textbf{0.81}} & {\textbf{0.81}} & 0.82
\end{tabular}
\caption{Comparison of ranking performance with \netl features and \lgnb 
features. Boldface indicates the better system between \netl and \lgnb 
(with an absolute difference $> 0.01$).}
\label{tab:cand-rank-results}
\end{table}

\section{Discussion}
\label{sec:discussion}

\begin{table}[t]
\footnotesize
\centering
\begin{tabular}{lc@{\;}cc@{\;}cc@{\;}cc@{\;}c}
\textbf{Test Domain} & \multicolumn{2}{c}{\textbf{\blogs}} & 
\multicolumn{2}{c}{\textbf{\books}} & \multicolumn{2}{c}{\textbf{\news}} 
& \multicolumn{2}{c}{\textbf{\pubmed}} \\
\hline
\textbf{All Features} & 2.00 & & 1.99 & & 2.02 & & 1.99 & \\
\hdashline
\textbf{$-$\lettertrigram} & 1.99 &\diff($-$.01)& 1.96 &\diff($-$.03)& 2.02 
&\diff($\pm$.00)& 1.93 &\diff($-$.06) \\
\textbf{$-$\pagerank} & 1.93 &\diff($-$.07)& 1.980 &\diff($-$.01)& 2.00 
&\diff($-$.02)& 2.00 &\diff($+$.01) \\
\textbf{$-$\topicoverlap} & 2.00 &\diff($\pm$.00)& 2.03 &\diff($+$.04)& 
2.04 &\diff($+$.02)& 1.95 &\diff($-$.04) \\
\textbf{$-$\wordnum} & 1.97 &\diff($-$.03)& 2.02 &\diff($+$.03)& 2.02 
&\diff($\pm$.00)& 2.00 &\diff($+$.01) \\
\end{tabular}
\caption{Feature ablation results based on in-domain top-1 average 
ratings.}
\label{tab:feat-ablation}
\end{table}

To better understand the contribution of each feature in \netl, we perform 
feature ablation tests (\tabref{feat-ablation}).  An interesting 
observation is that different features appear to have different impact 
depending on the domain. Looking at \blogs, we find that there is a 
considerable drop in top-1 average rating when we remove the 
\pagerank feature.  Similarly, ablating \lettertrigram appears to have a 
significant impact on \pubmed as well as some influence on \books.  
As far as \news is concerned, we observe feature ablation does not play a big role.
These observations indicate there is some degree of complementarity between 
these features, and that combining them produces robust and consistent performance across 
different domains.


Additionally, we explored using different numbers of topic terms when
computing topic and title relevance for candidate ranking (we tested 
using top-5/10/15/20 topic terms). In general,
we find that performance drops with the increase in topic terms.  We
also experiment with weighting each topic term with its word
probability.  We observed an improvement, although the difference is so
marginal that we omit the results from the paper. Lastly, we tried
computing relevance by first computing the centroid of topic terms
before computing the cosine similarity with a candidate title. Again, we
found little gain with this approach.

One feature type that we expect would have high utility is graph
connectivity over the graphical structure of the Wikipedia categories or
similar, along the lines of \newcite{hulpus+:2013}. We leave this to
future work. Methods based on keyphrase extraction such as
\newcite{zhao+:2011} are also potentially worth exploring, although it
remains to be seen whether notions such as ``interestingness'' benefit
topic label selection.


\section{Conclusion}
\label{sec:conclusion}

We propose a neural embedding approach to automatically label topics 
using Wikipedia titles. Our methodology combines document and word 
embeddings to select the most relevant labels for topics. Compare to a 
state-of-the-art competitor system, our model is simpler, more 
efficient, and achieves better results across a range of domains.

\bibliographystyle{acl}
\bibliography{strings,refs,papers}

\end{document}